\pdfoutput=1

\documentclass[11pt]{article}

\usepackage[final]{acl}

\usepackage{times}
\usepackage{latexsym}

\usepackage[T1]{fontenc}

\usepackage[utf8]{inputenc}

\usepackage{microtype}

\usepackage{inconsolata}

\usepackage{graphicx}
\usepackage{amsmath}
\newcommand{\run}[1]{\textcolor{black}{#1}}

\title{Detecting Mental Manipulation in Speech via \\Synthetic Multi‑Speaker Dialogue}

\author{
    \textbf{Run Chen$^{1\ast}$}, 
    \textbf{Wen Liang$^{1,2\ast}$}, 
    \textbf{Ziwei Gong$^1$}, 
    \textbf{Lin Ai$^1$},
    Julia Hirschberg$^1$
    \\
    $^1$Columbia University, USA~~~~~~
    $^2$Red Hat, USA
    \\
    \small{\texttt{\{runchen, sara.ziweigong, lin.ai, julia\}@cs.columbia.edu, wl2904@columbia.edu}}
   \\
   \small{$^\ast$Equal contributions.}
}

\begin{document}
\maketitle
\begin{abstract}
\textbf{Mental manipulation}, the strategic use of language to covertly influence or exploit others, is a newly emerging task in computational social reasoning. Prior work has focused exclusively on textual conversations, overlooking how manipulative tactics manifest in speech. We present the first study of mental manipulation detection in spoken dialogues, introducing a synthetic multi-speaker benchmark \textsc{SpeechMentalManip} that augments a text-based dataset with high-quality, voice-consistent Text-to-Speech rendered audio. Using few-shot large audio-language models and human annotation, we evaluate how modality affects detection accuracy and perception. Our results reveal that models exhibit high specificity but markedly lower recall on speech compared to text, suggesting sensitivity to missing acoustic or prosodic cues in training. Human raters show similar uncertainty in the audio setting, underscoring the inherent ambiguity of manipulative speech. Together, these findings highlight the need for modality-aware evaluation and safety alignment in multimodal dialogue systems.

\end{abstract}

\section{Introduction}

\begin{figure}
    \centering
    \includegraphics[width=\linewidth]{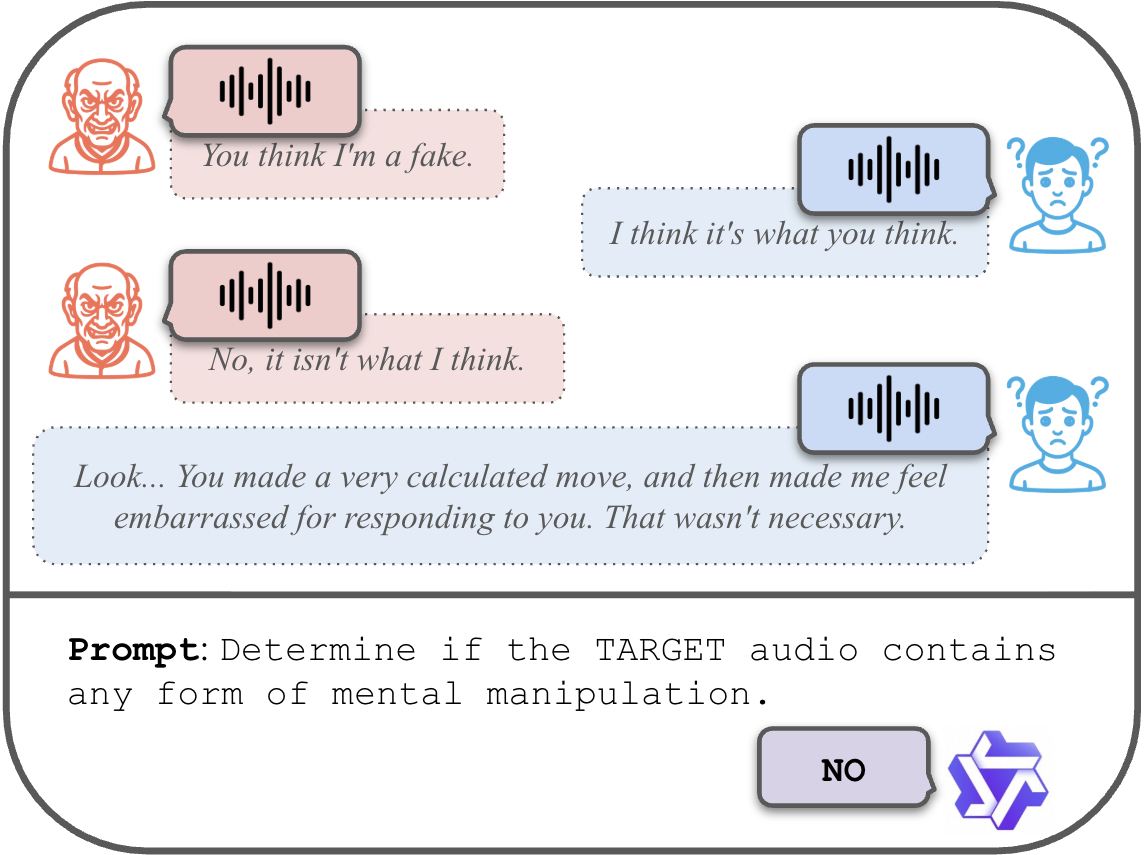}
    \caption{An example dialogue from the \textsc{SpeechMentalManip} dataset. The Qwen2.5 model is given the audio (transcript shown for clarity), but fails to detect manipulation.}
    \label{fig:example_dialogue}
\end{figure}

\textbf{Mental manipulation} refers to the covert use of tactics to steer another person's thoughts or emotions toward the manipulator's goals \citep{Barnhill2014}. Amplified by modern digital channels, its reach and precision have expanded from one-to-one interactions to broad, rapid dissemination, making targeted influence easier than ever \citep{Ienca2023}. The consequences are nontrivial: affected individuals often experience substantial psychological strain and mental-health burden \citep{Hamel2023}. Detecting such manipulation in dialogue remains a difficult challenge for computational social reasoning and safety, even for modern language models \citep{SimonFoley2011, gong-etal-2023-eliciting, wang2024mentalmanip, chen2025synthempathyscalableempathycorpus}. Beyond lexical content, real conversations rely on prosody, timing, and voice quality, which can reshape perceived intent. Understanding how these cues interact with linguistic strategies is essential for trustworthy multi-modal assistants. 
In parallel to research on manipulation safety, recent work in multimodal affect and emotion recognition has examined how emotion labels and modality cues interact in conversation \citep{gong-etal-2024-mapping} and identified methodological challenges in text–speech–vision integration \citep{wu2025merc-survey}.
These insights motivate our modality-aware design for manipulation detection in speech and connect to theory-of-mind style reasoning with LLMs in dialogue \citep{MoghaddamHoney2023,Chen2024,Strachan2024}.

Existing benchmarks for mental manipulation, however, focus almost entirely on text dialogues, leaving the role of prosody, tone, and delivery in manipulative speech largely unexplored. The \textsc{MentalManip} dataset formalizes manipulative presence and tactics in movie-style conversations, yet even strong LLMs struggle with text-only detection and attribution, with only modest gains from intent-aware prompting \citep{wang2024mentalmanip,ma2025iap}. However, audio-capable large multimodal models introduce distinct safety risks: speech is a sensitive attack surface and current systems can be brittle under adversarial or persuasive voice inputs \citep{yang2024audioachilles,peri2024speechguard,shen2024voicejailbreak}. These observations suggest that speech may indeed alter both the expression and detectability of manipulation, particularly for subtle tactics that require intent inference \citep{Kern2009,Lampron2024}.

To our knowledge, no existing benchmark connects manipulative content to \textbf{spoken} delivery, preventing systematic study of modality effects. 
We address this gap by introducing \textbf{\textsc{SpeechMentalManip}}\footnote{We release the dataset and code:
\url{https://github.com/runjchen/speech_mentalmanip}
}, a synthetic multi-speaker speech benchmark for mental manipulation (Figure \ref{fig:example_dialogue}).
The dataset extends \textsc{MentalManip} by rendering its textual dialogue transcripts into transcript-aligned, voice-consistent audio via a two-phase Text-to-Speech (TTS) pipeline (Figure \ref{fig:tts_pipeline}), thus enabling direct one-to-one comparisons between text and speech while explicitly probing the effects of prosodic cues. 
To examine how speech affects manipulation detection, we evaluate large pretrained audio-language models under few-shot learning \citep{Brown2020} and Chain-of-Thought reasoning setups \citep{Kojima2022}, juxtaposed with prior text-only results. We find that models show higher precision but markedly lower recall on audio, favoring conservative judgments that overlook subtle manipulative cues.

\run{
Following the observed model performance shift, human re-annotation of a representative subset further reveals lower cross-annotator agreement for audio than for text, highlighting modality-induced ambiguity and helping contextualize apparent model–label mismatches.}


In summary, our contributions are threefold:
(1) a new benchmark, \textsc{SpeechMentalManip}, that extends manipulation detection into speech;
(2) a evaluation of large audio-language model performance under few-shot and reasoning-based prompts; and
(3) a human re-annotation study revealing modality-driven ambiguity in manipulation perception.
Together, these establish the first systematic benchmark and analysis of mental manipulation in speech, emphasizing the need for modality-aware evaluation and alignment in multimodal dialogue safety.

\section{Related work}

\paragraph{Mental Manipulation in Dialogue}

Prior work on mental manipulation has focused primarily on the text modality. The \textsc{MentalManip} dataset introduces 4k movie-dialogue snippets with fine-grained labels for presence, technique, and targeted vulnerability, and shows that LLMs struggle on text-only detection and attribution \citep{wang2024mentalmanip}. 
Subsequent studies explore improvements through speaker intent-aware prompting in the Theory-of-Mind (ToM) style \cite{ma2025iap}, Chain-of-Thought (CoT) reasoning \cite{yang2024enhanceddetectionconversationalmental}, and a multi-task anti-curriculum distillation approach \cite{gao2025mentalmacenhancinglargelanguage}, aimed at enhancing interpretability and reduce false negatives over standard few-shot baselines.
Mental manipulation forms part of a broader class of social-reasoning and safety challenges in multimodal dialogue.

\begin{table}[t]
\centering
\small
\begin{tabular}{lcc}
\hline
\textbf{Technique} & \textbf{Count} & \textbf{\%} \\
\hline
Persuasion or Seduction & 607 & 25.87 \\
Shaming or Belittlement & 384 & 16.37 \\
Accusation              & 361 & 15.39 \\
Intimidation            & 321 & 13.68 \\
Rationalization         & 213 &  9.08 \\
Brandishing Anger       & 133 &  5.67 \\
Denial                  &  87 &  3.71 \\
Evasion                 &  83 &  3.54 \\
Playing Victim Role     &  69 &  2.94 \\
Feigning Innocence      &  58 &  2.47 \\
Playing Servant Role    &  30 &  1.28 \\
\hline
\end{tabular}
\caption{Distribution of ground-truth manipulation tactics across labeled instances in \textsc{MentalManip\_con}, the consensus subset with unanimous prior annotations.}
\label{tab:gt-tactics}
\end{table}

\paragraph{LMMs safety} Recent work on large multimodal models (LMMs) highlights unique safety failure modes in the audio route. 
Red-teaming studies show that audio is a sensitive attack surface for multimodal systems \citep{yang2024audioachilles}. Concurrently, \citet{peri2024speechguard} analyze adversarial robustness of speech-instruction language models and propose countermeasures, while \citet{shen2024voicejailbreak} demonstrates persuasive, story-driven “voice jailbreaks” against GPT-4o’s voice mode. These findings collectively motivate modality-specific evaluation and curation for manipulation detection in speech.

Despite growing awareness of these multimodal safety risks, there remains no benchmark that systematically links manipulative language to its spoken realization. In particular, the absence of controlled, transcript-aligned speech data makes it difficult to isolate how prosody, voice quality, and delivery influence the perception and detection of manipulation. Our work addresses this gap by augmenting the \textsc{MentalManip} dataset with high-fidelity, multi-speaker TTS renderings that preserve conversational structure and speaker identity, enabling direct comparison between text and audio.

\paragraph{Synthetic Speech}

Recent advances in expressive TTS have enabled natural-sounding, emotion-conditioned speech synthesis with controllable prosody and speaker identity. Systems leverage large-scale neural architectures and prompt-based conditioning to capture subtle affective and pragmatic cues such as tone, emphasis, and hesitation, extending beyond purely text-driven synthesis \cite{chen-etal-2024-emoknob}. Techniques such as prosody modeling and style transfer in Tacotron and VITS-based frameworks \citep{wang2018tacotron2,kim2021vits}, zero-/few-shot voice cloning \citep{jia2018transfer}, and expressive multi-style models \citep{wang2023valle,du2025cosyvoice, lyu2025build}. GPT-SoVITS\footnote{\url{https://github.com/RVC-Boss/GPT-SoVITS}} enables fine-grained control over speaker characteristics and emotional delivery, and expressive TTS has found growing applications in emotion-conditioned generation \cite{liang2025ECETTS}.
These advances make it feasible to generate multi-speaker, context-consistent dialogues with realistic prosody, which directly supports our study of manipulation detection in speech.

Most off-the-shelf TTS systems are optimized for single-speaker, single-turn synthesis; they lack key capabilities required for multi-turn dialogue synthesis: (i) robust multi-speaker dialogue rendering with stable identities across dozens of turns, (ii) precise control over timing and pauses needed to preserve conversational rhythm, or (iii) consistent prosodic coupling between adjacent turns. In practice, these issues lead to speaker drift, uneven loudness and pacing, and loss of turn-taking cues, which can confound downstream analysis of manipulation in speech. In addition, the streaming and batch modes of current TTS systems impose a quality-latency trade-off. To mitigate these issues, our approach (Figure~\ref{fig:tts_pipeline}) uses a deterministic speaker-voice mapping, synthesizes per-turn utterances, and composes them into a single continuous multi-speaker audio.

\section{Method}

\subsection{Dataset and Voice Pool}
Our study builds on the text-based dataset \textsc{MentalManip}\footnote{\url{https://github.com/audreycs/MentalManip/tree/main/mentalmanip_dataset}} \cite{wang2024mentalmanip}, which contains movie dialogue snippets derived from the Cornell Movie Dialogues corpus \cite{danescu-niculescu-mizil-lee-2011-chameleons} with fine-grained labels for manipulative presence and technique. 
Prior evaluation on such benchmark indicate that few-shot GPT-4 Turbo reaches 0.724 accuracy and a finetuned LLaMA-2-13B achieves 0.768 accuracy on the core detection task \citep{wang2024mentalmanip}. Incorporating intent-aware prompting in ToM style offers small but consistent gains, raising GPT-4-1106-Preview to 0.726 accuracy \cite{ma2025iap}.

\run{Rather than using original movie audio, we synthesize speech from the dialogue transcripts using TTS. The Cornell corpus provides dialogue scripts but does not include timestamps or aligned audio, making it infeasible to reliably extract corresponding speech segments without substantial manual effort. Moreover, many source movies are not freely redistributable, and licensing constraints preclude releasing aligned audio clips at scale. Using TTS allows us to generate transcript-aligned, shareable speech data with precise control over speaker identity and timing, enabling reproducible evaluation and direct comparison between text and audio modalities. This design prioritizes experimental control and accessibility over ecological realism, consistent with our goal of isolating modality effects.}

For our experiments, we construct the \textsc{SpeechMentalManip} dataset by synthesizing audio from the consensus split \textsc{MentalManip\_con}, which comprises 2,915 dialogue transcripts drawn from the original 4k dataset. This process yields 609 manipulative and 90 non-manipulative audio clips used for evaluation.


Each transcript is rendered into speech using a multi-speaker TTS pipeline (Figure~\ref{fig:tts_pipeline}), with consistent voice assignments per speaker to preserve identity and conversational coherence across turns. All results in this paper are reported on this audio-only evaluation set. To contextualize our experiments, Table~\ref{tab:gt-tactics} summarizes the ground-truth distribution of manipulation tactics aggregated over the \textsc{MentalManip\_con} split.

To generate the audio, we assign consistent, realistic voices to each speaker. As prior multimodal dialogue studies highlight that limited accent and demographic coverage can bias perception and annotation quality \cite{ace2025}, we vary speaker profiles and accents and later re-annotate labels in audio to account for these factors. We curate a fixed pool of six ElevenLabs voices spanning genders, ages, and accents (Table~\ref{tab:11labs-voices}); each speaker in a conversation is deterministically mapped to one voice to preserve speaker identity across turns.

\begin{table*}[t]
\centering
\small
\begin{tabular}{cccccc}
\hline
\textbf{Gender} & \textbf{Age} & \textbf{Language} & \textbf{Accent} & \textbf{Name} & \textbf{Voice ID} \\
\hline
F & Young adult   & English & American         & Ivanna -- Young \& Casual          & \texttt{yM93hbw8Qtvdma2wCnJG} \\
M & Young adult   & English & American         & Mark -- Natural Conversations      & \texttt{UgBBYS2sOqTuMpoF3BR0} \\
F & Mature adult  & English & American         & Amanda                              & \texttt{M6N6ldXhi5YNZyZSDe7k} \\
F & Middle-aged   & English & African American & Sassy Aerisita                      & \texttt{03vEurziQfq3V8WZhQvn} \\
M & Old           & English & American         & Grandpa Spuds Oxley                 & \texttt{NOpBlnGlnO9m6vDvFkFC} \\
F & Old           & English & American         & Grandma Muffin                      & \texttt{vFLqXa8bgbofGarf6fZh} \\
\hline
\end{tabular}
\caption{ElevenLabs voice pool used for multi-speaker rendering. Each speaker is mapped deterministically to one voice to preserve speaker identity across turns.}
\label{tab:11labs-voices}
\end{table*}

\subsection{Multi-Speaker TTS Audio Generation}
To isolate modality effects on manipulation detection from multi-turn conversations with diverse voice profiles, we require reproducible, voice-consistent, and transcript-aligned dialogue audio. Since end-to-end multi-speaker TTS remains limited for long conversational synthesis, we adopt a compose-from-turns strategy: (1) assign each speaker a fixed synthetic voice using ElevenLabs API \footnote{\url{https://elevenlabs.io/docs/api-reference/text-to-speech/convert}} deterministically and synthesize each utterance per turn; (2) concatenate these utterances into a single conversation clip with normalized loudness and controlled inter-turn silences (0.2s).  
This design preserves speaker identity, maintains alignment with the ground-truth (GT) transcripts from the \textsc{MentalManip\_con} dataset, and yields reproducible audio suitable for benchmarking and human evaluation. The scalable text-to-speech (TTS) workflow has two detailed phases (Figure~\ref{fig:tts_pipeline}):

\usetikzlibrary{positioning,arrows.meta,fit,calc}

\begin{figure}[t]
\centering
\scalebox{0.7}{%
\begin{tikzpicture}[
    font=\small,
    block/.style={rectangle, draw, fill=blue!20, text width=8.2em, text centered, rounded corners, minimum height=3.2em},
    big_block/.style={rectangle, draw, fill=gray!15, rounded corners=3pt, inner sep=8pt},
    tag/.style={rectangle, draw=black!40, fill=black!5, rounded corners=2pt, inner xsep=4pt, inner ysep=2pt},
    arrow/.style={-Latex, thick}
]

\node[block, text width=12em] (raw) {Raw Conversation Data};

\node[big_block, below=1.0cm of raw, minimum width=9.2cm, minimum height=7.6cm, anchor=north] (p1) {};
\node[font=\bfseries, anchor=north west] at ($(p1.north west)+(6pt,-4pt)$) {Phase 1: Turn-level audio generation};

\node[block, anchor=north] (meta) at ($(p1.north)+(0,-1.2cm)$) {Extract Metadata};
\node[block, below=14mm of meta] (assign) {Assign Voices to Speaker IDs};
\node[block, below=10mm of assign] (synth) {Generate Turn-Level Audio};

\node[tag, right=0.75cm of assign, align=center, text width=4em] (voices) {Predefined\\voice pool\\(Table~\ref{tab:11labs-voices})};

\node[big_block, below=1.2cm of p1.south, minimum width=9.2cm, minimum height=5.8cm, anchor=north] (p2) {};
\node[font=\bfseries, anchor=north west] at ($(p2.north west)+(6pt,-4pt)$) {Phase 2: Conversation reconstruction};

\node[block, anchor=north] (recon) at ($(p2.north)+(0,-1.2cm)$) {Reconstruct Conversations};

\node[tag, below left=7mm and 18mm of recon.south, align=center] (keeporder) {Preserve\\speaker order};
\node[tag, below=7mm of recon.south, align=center]               (ordert)    {Order by\\turn\_id};
\node[tag, below right=7mm and 18mm of recon.south, align=center](groupc)    {Group per\\conversation\_id};

\node[block, below=13mm of ordert, text width=16em] (compose) {Concatenate ordered segments\\into single conversation clip};

\draw[arrow] (raw) -- (meta);

\draw[arrow]
  (meta.south) -- node[midway, fill=white, inner sep=1pt]{\normalsize speaker\_id} (assign.north);

\coordinate (convleft) at ($(synth.west)+(-0.2,0)$);
\draw[arrow]
  ($(meta.south)+(-1.0,0)$)
  .. controls +(-0.2,-0.9) and +(-0.6,1.0) ..
  (convleft)
  node[pos=0.2, left=2pt, fill=white, inner sep=1pt]{\normalsize conversation\_id};

\coordinate (turnright) at ($(synth.east)+(0.2,0)$);
\draw[arrow]
  ($(meta.south)+(1.0,0)$)
  .. controls +(0.2,-0.9) and +(0.6,1.0) ..
  (turnright)
  node[pos=0.2, right=2pt, fill=white, inner sep=1pt]{\normalsize turn\_id};

\draw[arrow] (assign) -- (synth);

\draw[arrow] (voices.west) -- ++(-0.6,0) |- (assign.east);

\coordinate (reconTopIn) at ($(recon.north)+(0,0.8)$);
\draw[arrow, shorten >=1pt, shorten <=1pt]
  (synth.south) |- (reconTopIn) -- (recon.north);

\draw[arrow] (recon.south) -- (keeporder.north);
\draw[arrow] (recon.south) -- (ordert.north);
\draw[arrow] (recon.south) -- (groupc.north);

\draw[arrow] (keeporder.south) -- (compose.north -| keeporder.south);
\draw[arrow] (ordert.south)    -- (compose.north);
\draw[arrow] (groupc.south)    -- (compose.north -| groupc.south);

\end{tikzpicture}
}
\caption{Two-phase pipeline for TTS audio generation and conversational reconstruction.}
\label{fig:tts_pipeline}
\end{figure}
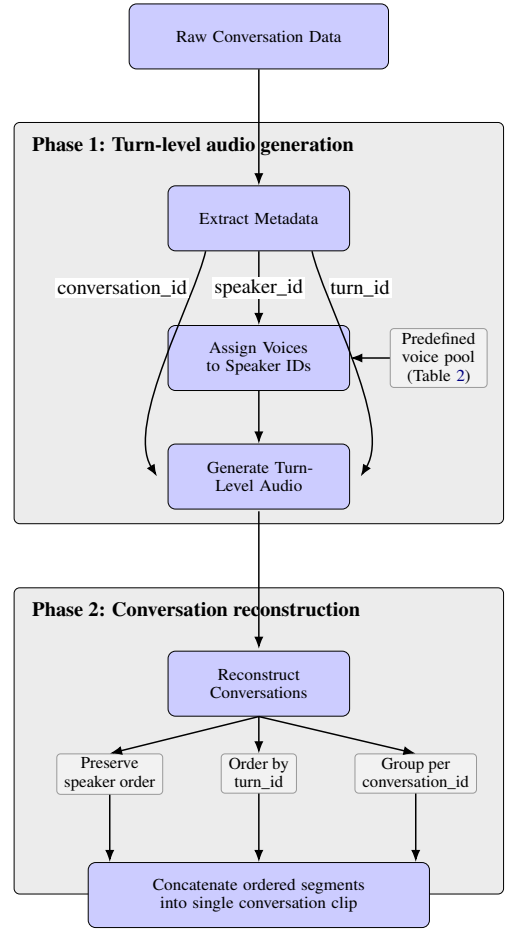

\paragraph{Phase 1: Turn-level audio generation.}
\begin{enumerate}
    \item Metadata extraction: For each raw conversation, we extract \textsc{speaker\_id, conversation\_id, and turn\_id}.
    \item Voice assignment: Each \textsc{speaker\_id} is deterministically assigned a distinct synthetic voice from a predefined pool (Table~\ref{tab:11labs-voices}) to ensure speaker consistency across all turns.
    \item Audio synthesis: We synthesize one audio file per utterance (turn) and store segments in a structured layout keyed by \textsc{conversation\_id} or \textsc{turn\_id} for downstream composition.
\end{enumerate}

\paragraph{Phase 2: Conversation reconstruction.}
\begin{enumerate}
    \item Dialogue composition: For each \textsc{conversation\_id}, we gather the synthesized utterances and order them by \textsc{turn\_id}, preserving the original speaker sequence.
    \item Output generation: We concatenate the ordered segments into a single composed clip per conversation, yielding a coherent multi-speaker recording suitable for audio-only evaluation.
\end{enumerate}

This two-phase process provides a flexible, efficient, and repeatable mechanism for converting text-based dialogues into lifelike, multi-voice synthetic conversations. It enables controlled studies of how emotions and acoustic cues in speech affect listener perception, engagement, and susceptibility to mental manipulation.

\subsection{Model Selection}
We use Qwen2.5-Omni-7B (Thinker-only) \footnote{\url{https://huggingface.co/Qwen/Qwen2.5-Omni-7B}} as our evaluation model due to its stable audio comprehension, balanced response behavior, and reliable adherence to constrained few-shot prompting. In preliminary trials, Qwen consistently ingests speech audio and follows constrained decoding and few-shot instructions without a systematic bias toward positive (manipulative) predictions. 
In contrast, other audio-language models we piloted, such as SALMONN \cite{tang2024salmonn} and Gemini-2.5-Pro \cite{comanici2025gemini25}, under their default system prompts and unconstrained decoding, frequently over-flag generic “violation/safety” cues (e.g. agitated prosody), yielding a persistent bias toward the manipulative label even on negative ground-truth clips. 
Because this systematic over-flagging prevents meaningful analysis, we focus on Qwen, which allows us to analyze modality effects, tactic distributions, and error patterns under controlled prompting conditions, without confounding manipulation inference with pervasive false positives driven by safety alignment mechanisms.


\section{Experiment Setup}
\subsection{Few-shot Detection Pipeline}
We run an audio-only batch evaluation pipeline to assess detection of mental manipulation and tactic attribution. Each query prompt is preceded by four labeled exemplars (two non-manipulative and two manipulative) that define the expected output format: a binary decision, a single best tactic, and one short supporting quote. Full prompts are detailed in Appendix~\ref{sec:prompt-pack}).

We formulate detection as a binary \textsc{YES}/\textsc{NO} task with A/B–constrained decoding. For each of five runs, we first apply this constraint; if it fails, we fall back to a single-token logit decision comparing the marginalized probabilities of the \textsc{YES} vs.\ \textsc{NO} verbalizers and predict \textsc{YES} iff $p(\text{YES})>p(\text{NO})$. This fallback captures the model’s immediate class preference while avoiding exposure/length biases from multi-token decoding and aligns with prompt-likelihood scoring. Moderate sampling is used only for the votes (temperature $=0.6$, top-$p=0.95$). The final clip label is the majority over the five run-level labels, following self-consistency sampling to improve robustness and accuracy \citep{wang2022selfconsistency}.

For clips predicted as manipulative (\textsc{YES}), we further infer the tactic label.
The full tactic inventory includes \{\emph{Accusation}, \emph{Brandishing Anger}, \emph{Denial, Evasion}, \emph{Feigning Innocence}, \emph{Intimidation}, \emph{Persuasion or Seduction}, \emph{Playing Servant Role}, \emph{Playing Victim Role},
\emph{Rationalization}, \emph{Shaming or Belittlement}, \emph{none}\}. We run five passes with
the same sampling as before (temperature $=0.6$, top-p$=0.95$) and select by majority vote. In each pass, tactics are scored by first-token probabilities; if the top option is \emph{none} or its margin over the runner-up is $<0.03$, we select the second-best. If the vote top-count is tied and the tie includes \emph{none}, we compute the mean first-token probability per tied label across votes and select a non-\emph{none} label only if it exceeds the mean probability of \emph{none} by $\geq 0.02$; otherwise we emit \emph{none}. This balances precision and recall while avoiding arbitrary tie resolution.

For any \textsc{YES} prediction, we require a single concise supporting quote for evidence and apply light post-processing (whitespace and quote normalization) without any semantic filtering or re-ranking; empty outputs trigger one retry with a shortened prompt.

\subsection{Evaluation Protocol and Metrics}\label{sec:eval-protocol}
We evaluate the speech manipulation detection at the clip level, treating \textsc{YES} (manipulative) as the positive class and \textsc{NO} (non-manipulative) as the negative class. Each clip undergoes five stochastic passes (temperature $=0.6$, top-p$=0.95$), and the final label is determined by majority vote.

To separate sensitivity from specificity, we compute confusion counts independently for the two composed-audio sets: GT$=\textsc{YES}$ and GT$=\textsc{NO}$ and report per-set accuracies 
(Table~\ref{tab:fewshot-cls-report}).

We analyze manipulative tactic attribution and evidence generation qualitatively to interpret model behavior. 
We summarize tactic distributions only among clips the model predicted \textsc{YES} within each ground-truth set. Percentages are taken with respect to the number of clips predicted \textsc{YES} in that set (e.g., 87 for GT $=\textsc{YES}$ and 16 for GT $=\textsc{NO}$), as shown in Tables~\ref{tab:pred-tactics-gt-yes} and \ref{tab:pred-tactics-gt-no}. This conditional analysis highlights which categories the model relies on when it asserts manipulation.
Similarly, each \textsc{YES} prediction is also paired with a short supporting quote (or brief paraphrase) after light normalization; these excerpts serve as interpretive context for understanding models rationale and error patterns.


\section{Audio-only Few-shot Detection Results}

We evaluate a five-pass, majority-vote pipeline on two composed-audio corpora: a manipulative set (GT{=}\textsc{YES}) and a non-manipulative set (GT{=}\textsc{NO}).
The model achieves 82.2\% accuracy on GT{=}\textsc{NO} and 34.8\% accuracy on GT{=}\textsc{YES} (Table~\ref{tab:fewshot-cls-report}); this indicates a sensitivity-specificity gap in which it avoids false alarms but under-detects many manipulative clips.

Examining tactic distributions (Table~\ref{tab:pred-tactics-gt-yes}), the true positive set (GT{=}\textsc{YES}, Pred{=}\textsc{YES}) concentrate on a small number of head classes: primarily \emph{Intimidation} (49.4\%) and \emph{Persuasion or Seduction} (29.9\%), while mid- and long-tail tactics present in the corpus (Table~\ref{tab:gt-tactics}) are rarely predicted. A similar pattern appears among false positives (GT{=}\textsc{NO}, Pred{=}\textsc{YES}) in Table~\ref{tab:pred-tactics-gt-no}, which are dominated by \emph{Persuasion or Seduction} (56.3\%) and \emph{Intimidation} (37.5\%). 
Together, these trends suggest a reliance on prosodic cues associated with arousal and valence, such as the acoustic pressure (e.g., loudness, sternness) of \emph{Intimidation} or the warmth of \emph{Persuasion}, resulting in a collapse toward these acoustically salient categories.

The modality mismatch probably exacerbates these effects: ground-truth tactics are transcript-based, while evaluation here is audio-only. Semantically defined tactics (e.g., \emph{Rationalization}, \emph{Denial}/\emph{Evasion}) may be weakly marked in prosody, while TTS delivery can amplify cues aligned with \emph{Intimidation} or \emph{Persuasion}. Combined with long-tailed class frequencies (e.g., \emph{Playing Servant Role} at 1.3\%) and overlapping definitions (e.g., \emph{Accusation} vs.\ \emph{Shaming}), the result is systematic under-detection of semantic tactics and over-reliance on a few dominant labels.

Notably, several clips labeled GT{=}\textsc{NO} nevertheless contain utterances that the model highlights as manipulative (Pred{=}\textsc{YES}), 
which illustrate points of ambiguity where perceived manipulative intent depends on context, delivery, and interpretation.
For example, the model surfaced evidence such as “Just as she starts feeling awful, you come up from behind and touch her neck.” (flagged as \emph{Intimidation}) and “I'm in love with you. How do you like that?” (flagged as \emph{Persuasion or Seduction}). We list four representative false positive cases with their predicted tactics and quoted spans in Appendix~\ref{app:qual-examples}. 
Because manipulation judgments are inherently subjective and context-dependent, the quoted spans should be interpreted as suggestive signals rather than definitive proof.

These apparent mismatches may arise from the model’s reliance on tactic name semantics, limited conversational context in short clips, or artifacts in the TTS delivery. These cases indicate residual label noise and motivate human re-annotation.

\begin{table}[t]
\centering
\small
\setlength{\tabcolsep}{7pt}
\begin{tabular}{lcccc}
\hline
\hline
\multicolumn{5}{c}{\textbf{Classification report}}\\
\hline
\textbf{Class} & \textbf{Precision} & \textbf{Recall} & \textbf{F1} & \textbf{N} \\
\hline
\textbf{GT{=}YES} & 0.845 & 0.348 & 0.493 & 250 \\
\textbf{GT{=}NO}  & 0.312 & 0.822 & 0.453 &  90 \\
\hline
\multicolumn{1}{l}{Macro avg}    & 0.578 & 0.585 & 0.473 & 340 \\
\multicolumn{1}{l}{Weighted avg} & 0.704 & 0.474 & 0.482 & 340 \\
\hline
\hline
\multicolumn{5}{c}{\textbf{Per-set accuracy}}\\
\hline
 & \textbf{Pred YES} & \textbf{Pred NO} & \textbf{Acc} & \textbf{N} \\
\hline
\textbf{GT{=}YES} & 87 & 163 & 0.348 & 250 \\
\textbf{GT{=}NO} & 16 &  74 & 0.822 &  90 \\
\hline
\hline
\end{tabular}
\caption{Consolidated results for the audio-only few-shot evaluation. Top: standard classification report over both sets combined. Bottom: per-set accuracies computed from the confusion counts. Supports (N) are GT counts (GT=YES: 250; GT=NO: 90).}
\label{tab:fewshot-cls-report}
\end{table}

\begin{table}[t]
\centering
\small
\begin{tabular}{lrr}
\hline
\textbf{Technique} & \textbf{Count} & \textbf{\%} \\
\hline
Intimidation            & 43 & 49.43 \\
Persuasion or Seduction & 26 & 29.89 \\
Shaming or Belittlement & 12 & 13.79 \\
Accusation              &  4 &  4.60 \\
Playing Servant Role    &  2 &  2.30 \\
\hline
\end{tabular}
\caption{Predicted tactic distribution within clips predicted \textsc{YES} for the GT=YES set (N${=}250$). Predicted \textsc{YES}$=87$, \textsc{NO}$=163$.}
\label{tab:pred-tactics-gt-yes}
\end{table}

\begin{table}[h]
\centering
\begin{tabular}{lrr}
\hline
\textbf{Technique} & \textbf{Count} & \textbf{\%} \\
\hline
Persuasion or Seduction &  9 & 56.25 \\
Intimidation            &  6 & 37.50 \\
Accusation              &  1 &  6.25 \\
\hline
\end{tabular}
\caption{Predicted tactic distribution within clips predicted \textsc{YES} for the GT=NO set (N${=}90$). Predicted \textsc{YES}$=16$, \textsc{NO}$=74$.}
\label{tab:pred-tactics-gt-no}
\end{table}

\section{Human Analysis of Modality-Induced Ambiguity}
\label{sec:why-recurate}

The preceding analysis reveals systematic mismatches between model predictions and the annotated ground truth, particularly in the speech modality. To better understand whether these divergences reflect model error, annotation ambiguity, or modality-induced perceptual differences, we conduct a targeted human analysis. The goal of this analysis is not to establish a definitive gold-standard label set, but to characterize how consistently humans perceive manipulative intent across text and speech. By examining inter-annotator agreement and cross-modality discrepancies, we contextualize the model behaviors observed above and assess the extent to which manipulation judgments are inherently subjective and modality dependent.


\subsection{Annotation Method}
We prepared 100 source conversations, each rendered in two modality-specific items: text-only (transcript) and audio-only (composed multi-speaker TTS). Each modality was annotated independently to prevent cross-modal leakage.

\run{Annotators were student volunteers fluent in English who completed the task independently and had no access to model predictions or ground-truth labels.} Eight annotators participated in total. Items were organized into ten batches per modality (IDs 0–9). \run{Each annotator was assigned one text batch and one audio batch in randomized order and was provided with definitions of mental manipulation and annotation guidelines (see Appendix \ref{app:annotation-interface} for interface details and instructions)}. This design ensured multiple independent judgments per item in each modality, with approximately 20–50\% overlap across annotators to support cross-validation.
The labeling task in this re-curation phase was intentionally narrow: annotators provided only the binary manipulative label \(\{ \textsc{YES}, \textsc{NO} \}\) for the given modality. Tactic labels were intentionally de-prioritized and not collected here.

To maintain data quality, we checked each item for annotation completeness and consistency. These checks included verification of valid class membership in \(\{ \textsc{YES}, \textsc{NO} \}\), batch integrity, and annotator–item uniqueness. Evidence quotes were not required at this stage.

After collection, labels were aggregated by majority vote within each item–modality pair. Let an item receive \(k\) votes \(y_i \in \{0,1\}\) with \(1=\textsc{YES}\). The final label \(\hat{y}\) is
\[
\hat{y} =
\begin{cases}
1, & \text{if } \sum_{i=1}^{k} y_i \ge \left\lceil \frac{k}{2} \right\rceil \\
0, & \text{if } \sum_{i=1}^{k} y_i \le \left\lfloor \frac{k}{2} \right\rfloor
\end{cases}
\]
and items with \(\sum_{i=1}^{k} y_i = \frac{k}{2}\) (a tie) were marked \textsc{UNRESOLVED} and routed to adjudication.

For adjudication, tied or low-confidence items were reviewed by two rotating annotators who were not in the original voting set for that item. They examined only the modality under review and issued a consensus \(\textsc{YES}/\textsc{NO}\). If consensus could not be reached, a third adjudicator served as a tie-breaker.

Finally, for each item and modality we recorded the resulting binary label, the vote histogram \((\#\textsc{YES}, \#\textsc{NO})\), the adjudication status, and annotator counts per item. After all batches closed, inter-annotator agreement metrics, including Cohen's Kappa \cite{cohen1960kappa}, Fleiss’s Kappa \cite{fleiss1971kappa} and Krippendorff’s Alpha \cite{krippendorff2004content}, were computed separately for each modality.

\subsection{Annotation Results and Discussions}
Given the inherently subjective nature of mental manipulation, we compare model performance with human annotations on the same tasks. We observe that human judgments occasionally diverge from the original task labels, and such discrepancies are more pronounced in the speech modality.

\begin{figure}[t]
    \centering
    \includegraphics[width=0.9\linewidth]{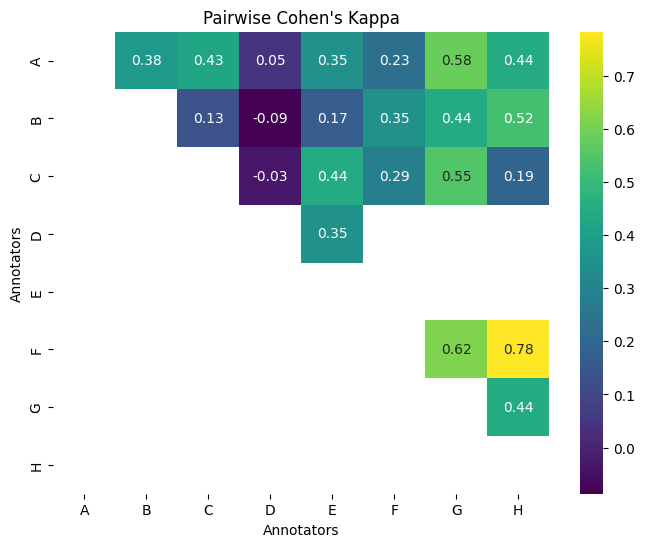}
    \caption{Pair-wise Cohen's Kappa between Human Annotators for \textit{Text} modality}
    \label{fig:cohens_kappa_text}
\end{figure}

We collect human judgments on the dialogues presented in either text or TTS audio modalities. We calculate the inter-annotator agreement represented in pair-wise Cohen’s Kappa, Fleiss’s Kappa and Krippendorff’s Alpha.
As pairwise Cohen's Kappa vary by a large degree (Figure \ref{fig:cohens_kappa_text}), we focus on the annotators with higher agreement. The high agreement group (annotators B, F, G, H) for text has Krippendorff's alpha of 0.526 and Fleiss's Kappa of 0.513. These values are slightly lower than the Fleiss’s Kappa of 0.596 reported in the original \textsc{MentalManip} dataset \cite{wang2024mentalmanip}. 

In audio modality, the high agreement group (annotators B, C, F, H) for text has Krippendorff's alpha of 0.422 and Fleiss's Kappa of 0.514. We observe that some annotators achieve higher agreement on the text modality but not necessarily on audio, suggesting that modality introduces additional variability in how manipulation cues are perceived.

\begin{figure}[t]
    \centering
    \includegraphics[width=0.9\linewidth]{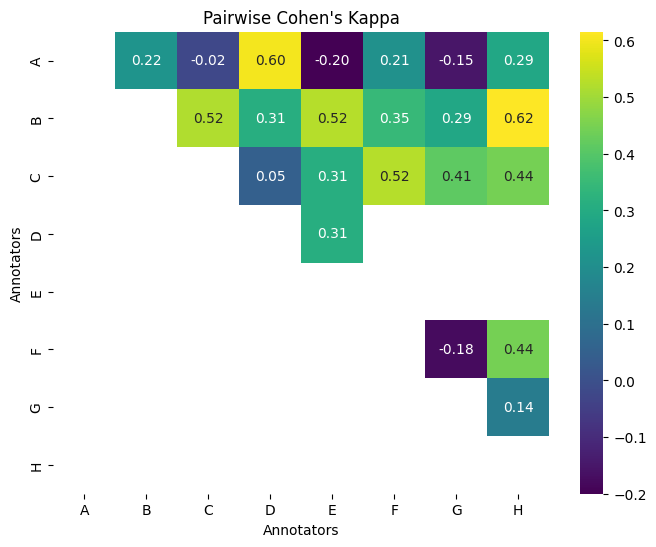}
    \caption{Pair-wise Cohen's Kappa between Human Annotators for \textit{Audio} modality}
    \label{fig:cohens_kappa_audio}
\end{figure}

Using majority voting over 100 re-annotated samples, we find that our labels align with the original \textsc{MentalManip} annotations at 0.72 agreement for text and 0.56 for audio, suggesting notably lower consistency in the speech modality. This discrepancy indicates that identifying mental manipulation from speech cues is inherently more ambiguous, probably due to prosodic and contextual subtleties that were underrepresented or inconsistently interpreted in the original dataset. The lower audio agreement also suggests that the original labels may not fully capture the nuanced intentions conveyed through tone, hesitation, or emphasis, which are features that often alter perceived manipulation. 

While we do not re-score model performance against the re-annotated labels in this work, we expect that using modality-faithful, audio-first annotations would reduce apparent false positives and increase measured recall, particularly for borderline cases where human judgments diverge from transcript-based labels. In this sense, some model errors observed under the original labels likely reflect annotation mismatch rather than incorrect inference.
Additionally, harder detectability in speech does not necessarily imply greater harm; it may reflect both weaker manipulation delivery and increased perceptual ambiguity, motivating future work that disentangles detectability from downstream listener impact.

\begin{table}[t]
\centering
\small
\begin{tabular}{|c|cc|cc|}
\hline
\textbf{Annotations} & \multicolumn{2}{c|}{\textbf{Text}} & \multicolumn{2}{c|}{\textbf{Audio}} \\
\hline
\textbf{\textsc{MentalManip}} & YES & NO & YES & NO \\
\hline
YES & 31 & 19 & 28 & 22 \\
NO & 9 & 41 & 22 & 28 \\
\hline
\end{tabular}
\caption{Agreement between the original \textsc{MentalManip} labels and our re-annotations for 100 samples.}
\label{tab:agreement}
\end{table}

\section{Conclusion}
We introduce the first benchmark \textsc{SpeechMentalManip} for detecting mental manipulation in speech by augmenting the text-based dataset with high-quality, voice-consistent TTS–rendered dialogues. This synthetic multi-speaker extension enables direct comparison between text and audio modalities while systematically examining how prosodic cues affect manipulative intent detection. Our experiments show that audio representations make the task substantially more challenging: both humans and models exhibit lower agreement and accuracy when manipulation must be inferred from speech rather than text. These findings highlight that mental manipulation is not only a difficult computational task but also an inherently subjective phenomenon, shaped by tone, delivery, and context. 

\textbf{Future work} will expand this benchmark toward more diverse voices and natural speech, refine theoretical definitions of manipulation, and explore modeling strategies that explicitly account for subjectivity and multimodal ambiguity, as explored in other social-pragmatic phenomena (e.g., empathy \citep{srikanth2025mixed}). As perception of manipulation can vary widely across individuals and contexts, clearer theoretical grounding is essential to ensure consistency in both human judgments and machine predictions. 
We will use the re-annotated audio-first labels as an alternative evaluation set to quantify how modality-faithful annotation reshapes precision–recall trade-offs and tactic attribution. 
We hope this work lays a foundation for developing safer, more socially aware dialogue systems that can reason about manipulative intent across modalities.


\section*{Ethical Statement}
Our findings show that manipulative intent is harder to consistently detect in spoken dialogue than in text, for both models and human annotators. This result should not be interpreted as evidence that speech-based manipulation is inherently more harmful or effective. An alternative interpretation is that current text-to-speech systems may not yet convey manipulative strategies with sufficient fidelity for them to reliably influence listeners, and that poorly realized manipulation may lose its persuasive impact. Importantly, our study examines detectability and agreement, not the effectiveness or outcomes of manipulation on human behavior. As such, reduced detectability should not be equated with increased harm. We emphasize the need for future work that jointly examines manipulation generation, perception, detectability, and listener impact to more fully assess ethical and safety implications.

\section*{Limitations}
Our task involves inherently subjective judgments, as perceptions of mental manipulation can vary across annotators and contexts. While we curate samples from the consensus set, the re-annotated samples may capture only a subset of manipulative strategies represented in the original dataset, limiting generalizability. 

In addition, our use of text-to-speech (TTS) synthesis for some audio stimuli may not fully reflect the richness and variability of natural human speech, potentially affecting both human and model interpretation.
Our synthetic dialogues are generated on a turn-by-turn basis and therefore do not capture overlapping speech, interruptions, or backchanneling commonly observed in natural conversation. This design choice prioritizes experimental control: overlapping speech remains challenging for current audio-language models and can introduce confounds related to speech separation, diarization, and acoustic comprehension. 
As our goal is to isolate how prosodic cues and delivery affect manipulation reasoning, rather than to stress-test low-level audio robustness, we intentionally evaluate models under clean, non-overlapping conditions.
Despite these limitations, we do not claim that synthetic speech faithfully represents natural manipulative behavior, but to provide a controlled testbed for isolating modality effects. By rendering transcript-aligned speech with consistent speaker identities and minimized acoustic confounds, we can probe how audio-language models and humans interpret manipulative intent when lexical content is held fixed, an analysis that would be difficult to conduct with in-the-wild recordings.
Incorporating statistically generated overlap (e.g., via Behavior-SD style simulation) represents an important direction for future work, enabling evaluation under more ecologically realistic conversational dynamics once baseline behaviors are established.


Finally, our evaluation relied on a single audio-language model (Qwen2.5-Omni) and a few-shot prompting strategy that did not include explicit definitions of manipulation tactics. While this choice established a stable baseline and tested the model's inherent semantic understanding, it leaves open the question of whether definition-augmented prompting or alternative architectures would yield different sensitivity patterns. Expanding the benchmark to a broader suite of models and prompt strategies remains a critical direction for future work.

\section*{Acknowledgments}


This work was supported in part by the funds provided by the National Science Foundation and by DoD OUSD (R\&E) under Cooperative Agreement PHY-2229929 (The NSF AI Institute for Artificial and Natural Intelligence). The views, opinions and/or findings expressed are those of the authors and should not be interpreted as representing the official views or policies of the National Science Foundation or the U.S. Government.

\bibliography{custom}
\clearpage
\appendix


\section{Prompts in Experiments}\label{sec:prompt-pack}
Our prompting setup follows prior work on MENTALMANIP~\citep{wang2024mentalmanip} for consistency. 
\begin{promptboxA}{Prompt (System + Few-shot + Tasks)}
\promptstart
\textbf{SYSTEM:}\pline
You are Qwen, a virtual human developed by the Qwen Team, Alibaba Group, capable of perceiving auditory and visual inputs, as well as generating text and speech.\par

\textbf{USER:}\pline
FEW-SHOT EXAMPLES (not the target).\pline
Use labels only for calibration.\par

Example (NOT manipulative):\pline
[audio: \textless EX1\_NO\_AUDIO\textgreater]\pline
Label: No\par

Example (NOT manipulative):\pline
[audio: \textless EX2\_NO\_AUDIO\textgreater]\pline
Label: No\par

Example (manipulative):\pline
[audio: \textless EX3\_YES\_AUDIO\textgreater]\pline
Label: Yes\par

Example (manipulative):\pline
[audio: \textless EX4\_YES\_AUDIO\textgreater]\pline
Label: Yes\par

Now analyze the TARGET audio below. Do NOT relabel examples above.\par
\textbf{TARGET audio:}\pline
[audio: \textless TARGET\_AUDIO\textgreater]\par

\textbf{Binary (YES/NO)}\pline
Task: Determine if the TARGET audio contains any form of mental manipulation.\pline
Choose exactly one option and output ONLY the letter on the first line.\pline
(A) YES — clear evidence of manipulation\pline
(B) NO  — otherwise\pline
Answer (A or B):\par

\textbf{YES/NO Fallback (single token)}\pline
[If the A/B letter is not produced, answer this instead:]\pline
You saw FEW-SHOT examples. For the TARGET only, answer YES or NO.\pline
Answer:\par

\textbf{Tactic (single label)}\pline
You saw FEW-SHOT examples above. Now classify the TARGET audio.\pline
Task: Choose the single best manipulation tactic for the TARGET from the list below.\pline
If there is no clear manipulation, choose `none'.\pline
Options: Accusation, Brandishing Anger, Denial, Evasion, Feigning Innocence,\pline
Intimidation, Persuasion or Seduction, Playing Servant Role,\pline
Playing Victim Role, Rationalization, Shaming or Belittlement, none\pline
Rule: Answer with exactly one option word from the list, nothing else.\pline
Answer:\par

\textbf{Evidence}\pline
You saw FEW-SHOT examples above. For the TARGET audio only, output ONE short quote\pline
(or paraphrase) that supports the given tactic (\(\leq\)12 words is ideal but not required).\pline
Tactic: \{tactic\}\pline
CRITICAL RULES:\pline
1) Output ONLY the quote/paraphrase wrapped in double quotes.\pline
2) No prefixes like Reason:, Example:, Description:, Source:, Tactic:.\pline
Answer:\par

\textbf{Evidence Retry}\pline
[If the evidence answer is empty or malformed, use this:]\pline
Output a quote from the TARGET in double quotes. Nothing else.\pline
Tactic: \{tactic\}\pline
Answer:
\end{promptboxA}

\section{Additional Qualitative Examples}\label{app:qual-examples}
We highlight the subjectivity and nuance of the mental manipulation task through several misaligned case studies. In the absence of an explicit victim response, ground-truth labels in the dataset often default to non-manipulative, whereas LLMs tend to interpret the potential manipulator’s utterance (typically the final turn) as evidence of manipulation.

\begin{promptboxWarn}{GT=\textsc{NO} \textrightarrow{} Pred=\textsc{YES}: Full transcripts with model evidence}
\promptstart

\textbf{Case 1}\pline
\textbf{Conversation ID:} \texttt{85514533}\pline
\textbf{GT:} \textsc{NO} \quad \textbf{Pred./Tactic:} \textsc{YES} / Intimidation\par
\textbf{Transcript:}\pline
\emph{Person1:} Howdy Pouty.\pline
\emph{Person2:} I was pretty confident that I was going to blow it with Talia, but I must say, I outdid myself.\pline
\emph{Person1:} She’s still pissed at me and took it out on you. We should have taken it slower. It’s hard to operate in the woods. Much easier in, like a club. Tell the girl you’ve got to go do something, leave her view, take way too long until she is worried that you’re not coming back. Just as she starts feeling awful, you come up from behind and touch her neck...\pline
\emph{Person2:} You are the prince of the darkness.\par
\textbf{Model evidence:}\pline
``Just as she starts feeling awful, you come up from behind and touch her neck.''\par

\medskip\noindent\rule{\linewidth}{0.4pt}\medskip

\textbf{Case 2}\pline
\textbf{Conversation ID:} \texttt{85514499}\pline
\textbf{GT:} \textsc{NO} \quad \textbf{Pred./Tactic:} \textsc{YES} / Persuasion or Seduction\par
\textbf{Transcript:}\pline
\emph{Person1:} You were quick enough to get Tom’s help when...\pline
\emph{Person2:} Yes, yes. I know. Right. And if it had gone well for me tonight, maybe I’d be keeping quiet about all this... I grant you everything but give me this... he does personify everything you’ve been fighting against... And I’m in love with you. How do you like that? — I buried the lead.\par
\textbf{Model evidence:}\pline
``I’m in love with you. How do you like that?''\par

\medskip\noindent\rule{\linewidth}{0.4pt}\medskip

\textbf{Case 3}\pline
\textbf{Conversation ID:} \texttt{85514451}\pline
\textbf{GT:} \textsc{NO} \quad \textbf{Pred./Tactic:} \textsc{YES} / Persuasion or Seduction\par
\textbf{Transcript:}\pline
\emph{Person1:} Yes, what?\pline
\emph{Person2:} Don’t answer me. Say what I say.\par
\textbf{Model evidence:}\pline
``Yes, what? Don’t answer me, say what I say.''\par

\medskip\noindent\rule{\linewidth}{0.4pt}\medskip

\textbf{Case 4}\pline
\textbf{Conversation ID:} \texttt{85514570}\pline
\textbf{GT:} \textsc{NO} \quad \textbf{Pred./Tactic:} \textsc{YES} / Persuasion or Seduction\par
\textbf{Transcript:}\pline
\emph{Person1:} Oh no, not you again.\pline
\emph{Person2:} What an adorable hat.\pline
\emph{Person1:} They think I have a concussion.\pline
\emph{Person2:} And you think you’re in love.\pline
\emph{Person1:} I know it.\pline
\emph{Person2:} This ``love'' of yours will soon wear off. I gave you a temporary love potion.\pline
\emph{Person1:} Why should I believe you?\pline
\emph{Person2:} It’s the truth. In twenty-four hours you’ll forget all about that girl.\par
\textbf{Model evidence:}\pline
``In twenty-four hours you’ll forget all about that girl.''\par
\end{promptboxWarn}

\section{Human Annotation Details}
\label{app:annotation-interface}

\begin{figure}[h]
    \centering
    \includegraphics[width=\linewidth]{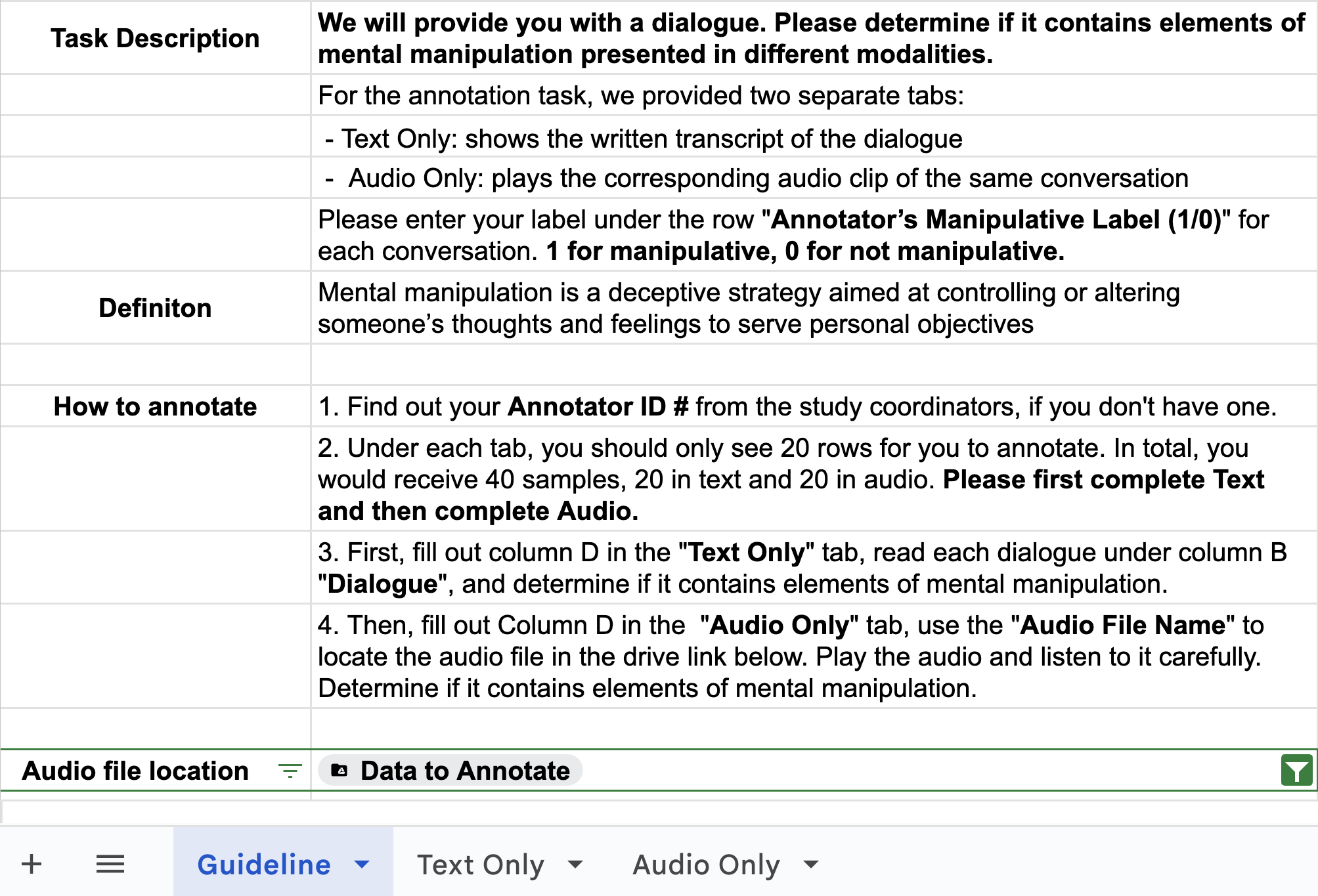}
    \caption{Annotation interface. Annotators first reviewed task instructions and the definition of mental manipulation (Guideline tab), then labeled the same dialogue under text-only and audio-only conditions in separate tabs.}
    \label{fig:annotation-interface}
\end{figure}

Figure~\ref{fig:annotation-interface} illustrates the annotation interface used in our human analysis. Annotators were provided with three tabs per assignment. The first tab presented task instructions and the shared definition of mental manipulation. The remaining two tabs each contained a single dialogue instance shown in one modality only, either Text Only (written transcript) or Audio Only (corresponding speech clip). 

In the Text Only tab, annotators saw the full written transcript of the conversation directly in the spreadsheet. 

In the Audio Only tab, annotators were given a link to the corresponding audio file hosted on Google Drive and were instructed to listen to the recording to make their judgment; no transcript was provided in the audio condition. 

The order of the text and audio tabs was randomized across annotators to control for order effects. Annotators assigned a binary label (0/1) indicating the presence or absence of mental manipulation independently for each modality, without access to tactic labels or model predictions.

\end{document}